# A Hybrid YOLOv5–SSD IoT-Based Animal Detection System for Durian Plantation Protection

Anis Suttan Shahrir [1] , Zakiah Ayop [2], Syarulnaziah Anawar [2] and Norulzahrah Mohd Zainudin [3]

[1] Menara Mesiniaga 1A, Jalan SS 16/1, 47500 Subang Jaya, Selangor, Malaysia
[2] Center for Advanced Computing Technology (C-ACT), Fakulti Teknologi Maklumat dan Komunikasi,Universiti Teknikal Malaysia Melaka (UTeM), 76100, Melaka, Malaysia
[3]Jabatan Sains Komputer, Fakulti Sains dan Teknologi Pertahanan, Universiti Pertahanan Nasional Malaysia (UPNM), 57000 Kuala Lumpur, Malaysia

## ABSTRACT

*Durian plantation suffers from animal intrusions that cause crop damage and financial loss. The traditional farming practices prove ineffective due to the unavailability of monitoring without human intervention. The fast growth of machine learning and Internet of Things (IoT) technology has led to new ways to detect animals. However, current systems are limited by dependence on single object detection algorithms, less accessible notification platforms, and limited deterrent mechanisms. This research suggests an IoT-enabled animal detection system for durian crops. The system integrates YOLOv5 and SSD object detection algorithms to improve detection accuracy. The system provides real-time monitoring, with detected intrusions automatically reported to farmers via Telegram notifications for rapid response. An automated sound mechanism (e.g., tiger roar) is triggered once the animal is detected. The YOLO+SSD model achieved accuracy rates of elephant, boar, and monkey at 90%, 85% and 70%, respectively. The system shows the highest accuracy in daytime and decreases at night, regardless of whether the image is still or a video. Overall, this study contributes a comprehensive and practical framework that combines detection, notification, and deterrence, paving the way for future innovations in automated farming solutions.*

## 1. INTRODUCTION

Durian cultivation in Malaysia is a significant and thriving industry, as Malaysia is one of the world's largest producers and exporters of durians. However, the environmental conditions in the plantation areas are still considered to be insecure from wildlife and animal threats that result in security and productivity concerns. According to The Malaysian Insight, an incident where wild elephants damaged crops has caused farmers to lose almost MYR200,000 despite various efforts to scare the herd away [1].

Animal detection system is relatively new in computer vision and the Internet of Things (IoT). In the past, people relied on traditional methods like tripwires or pressure pads to detect movement in a particular area. However, with the powerful cameras and advanced software, animal detection has become much more improved. Their ability to provide enhanced security, convenient access control, and compatible integration with other smart devices makes them essential to modern security systems.

The contributions of this study are as follows:

- Hybrid Detection Framework – Integration of YOLOv5 and SSD algorithms within a CNN-based IoT architecture to enhance detection accuracy.
- Real-Time Monitoring – Deployment of a camera-based IoT system capable of continuous surveillance and instant animal intrusion detection.
- Accessible Notification System – Direct integration with Telegram messaging to deliver timely alerts to farmers, overcoming reliance on less widely used platforms.
- Deterrent Mechanism – Implementation of a sound-based deterrent (tiger roar) to repel animals.
- Application in Agriculture – A practical case study focused on crop plantations.

This paper is organized as follows. Section 2 discusses the research background in Animal Detection and CNN algorithms. Section 3 introduces the proposed system methodology. Section 4 outlines the discussion of testing and results, and section 5 concludes the overall of this study.

## 2. RELATED WORK

### 2.1. Animal Detection System

The traditional method of animal detection involves direct human observation and the use of specialized tools and techniques. This includes visually scanning an area for animals, examining tracks and traces they leave behind, setting up trail cameras for remote monitoring, using acoustic devices to capture animal sounds, and employing methods for direct examination through trail camera ping and capture. These approaches have yielded valuable insights into animal behaviours and ecology but can be labour-intensive, time-consuming, and limited in coverage and scalability. Modern technologies are increasingly integrated with traditional methods to enhance animal detection and monitoring capabilities [2].

Building on this shift, Yusoff et al. [3] developed Pro Durian, a mobile application that uses Deep Neural Network (DNN) to detect animal with sound alert features. This application provides information on durian farm threats utilizing simple and minimal user interface for easy navigation. However, the application needs continuous human intervention as the farmer needs to bring their phone to the plantation area and open the camera on the application to capture the animal intrusion for image recognition process.

Megalingam et al. [4] design animal detection consisting of four junction boxes deploy as detection sensors. Each boxes contain a Light-Dependent Resistor (LDR) sensor used for detection, a XBee and GSM for communication and laser pointer to light up the area. However, designing sensor coverage is challenging in real life [5] due to performance light degradation as the distance increase. In addition, Xue et al. [6] researched an intrusion detection system using Ultra-Wide Band (UWB) technology in a wireless sensor network. The UWB signals will produce different signals due to varying object target. These signals are collected and classified using CNN algorithm to classify as human, animal or null. While this research contributes to CNN studies, the system is unable to differentiate between a threat or non-threat animal.

To address automation with limited human intervention, Panda et al. [7] proposed an IoT system where upon detection by ultrasonic sensors planted on the edge of the farm, a line robot carrying an ESP32 camera moves forward and capture images whenever motion is detected by the motion sensor installed in the robot. The image is forwarded to Blynk application and alerted to the farmer. The buzzer is activated to prevent animal intrusions. However, this model did not utilize image recognition as it is difficult to distinguish between a threat or non-threat animal.

To enhance detection accuracy, Sathesh et al. [5] employ YOLO algorithm to train animal datasets and build detection model. When the image match to the predefined animal trained in the model, the camera captures the image and uploads it to the server. Subsequently, a buzzer is turned on when the object match. The image captured will be emailed to the farmer. In addition, Anuvind et al. [8] developed a real-time animal detection system that utilize image and audio processing to automatically detect the presence of wild animals. The system employs YOLO4 for image recognition and Multilayer Perceptron (MLP) for audio recognition. When an intrusion is detected, the web-based system will send an alert message to relevant authorities. However, integrating two main processes to detect the presence of an intrusion may require a powerful processor for the system to perform smoothly without encountering any buffer while processing the image and audio of the detected animals. Similarly, Mamat et al. [9] proposed YOLOv5 model to detect four specific animals that commonly intrude farming areas. The implementation of the YOLOv5 demonstrated impressive results, achieving nearly 94% mean Average Precision (mAP) in detecting animal intrusion. With more enhancement within the system for a more user-friendly application, this system may have the potential as a reliable solution for farmers facing this issue.

## 2.2. Object Detection Algorithm

Animal detection systems rely heavily on object detection algorithms to function. Key algorithms like Region-based Convolutional Neural Networks (RCNN), Faster RCNN, Single Shot Detector (SSD), and You Only Look Once (YOLO) have been at the forefront of this change.

A Convolutional Neural Network (CNN) is a type of supervised multilayer neural network where the convolutional and pooling layers play crucial roles in feature extraction [10]. The output is passed to a fully connected layers and a classifier, which is responsible for classifying the samples. The features extracted by the convolutional layer are transferred through subsequent layers, culminating in the final output [11], [12]. Despite this process, there is often a discrepancy between the actual and ideal outputs. This error is addressed using the backpropagation (BP) algorithm, which transmits the error back through the network. By adjusting the network parameters based on this feedback, CNN can learn and optimize the data characteristics that enhance classification accuracy, thus refining the overall model [6].

Girshick et al. [13] introduced Region-based CNN (R-CNN) that consists of four parts: generating region proposals, feature extraction using CNN and classifying proposed regions with SVM and predicting bounding box using linear regression model. R-CNN produced high accuracy on the expense of costly computational power for embedded systems and high-end machines while too slow for real-time applications [14]. Fast R-CNN [15] rectifies the shortcomings of CNN by restructuring the first two phases and combining the last two phases. Fast R-CNN is twenty times faster than R-CNN because it uses a CNN to process a single picture rather than thousands, the region proposal is done over a low-dimensional feature map, and the prediction is combined. However, Fast R-CNN still consumes several seconds on average equipment to detect objects in one image despite the computational gain. Faster R-CNN [16] by Ren, et al. removes the selective search altogether, instead using a region proposal network (RPN), a learnable CNN that does not need inference and with higher speed with the standard hardware, making it possible to use in real-time object detection. The computational cost prohibits a single usage of this model as the most suitable for the on-board application, but suggested used alongside another technique with higher speed to complete the gaps between Faster R-CNN inferences.

The Single Shot Detector (SSD) was developed by Liu et al. [14] represents both faster and much more precise single-shot detector across many categories. The SSD is comparable to other slower methods that explicitly propose regions and perform pooling (like Faster R-CNN). Using small convolutional filters on feature maps, SSD primary function will forecast category scores and box offsets on a fixed set of default bounding boxes. Predictions are explicitly separated by aspect ratio, and varying scales are generated from multi-scale feature maps to achieve high detection accuracy.

The You Only Look Once (YOLO) algorithm is renowned to provide efficiency in the detection of targets. YOLO was founded by Redmon et al. in 2015 [17] and has since then come in several versions, up to YOLO v8. In the YOLO V5, the whole image goes through a single convolutional neural network and divides it into grids with the probability of the bounding boxes and the classes. The YOLO simplifies the detection procedure by reformulating the task of objection identification as a regressive operation removing the necessity of complex pipelines and increasing consistency performance in the real-time aspect [18], [19].

R-CNN and its subsequent versions have high detection accuracy because they use networks that put forward potential object regions to be analysed in detail. Faster R-CNN optimizes this process by using region proposals within the network and therefore, makes it faster [20]. SSD gives a suitable trade-off between speed and accuracy by identifying object borders and classes within a single process. YOLO is fast, and it is thus most suitable in a real-time application, albeit with a minor drop in precision compared to the RCNN-based approaches [21].

Despite the availability of YOLOv8, YOLOv5 is acceptable for most resource-limited IoT applications due to its precision, speed, and low computational overhead. YOLOv5 outperformed YOLOv8 in real-time panel defect detection with a faster inference speed (~7.1 ms/image) and higher accuracy (94.1%) [22]. In wildfire and smoke detection studies, YOLOv5s outperformed YOLOv8 in precision, recall, F1-score, and mAP@50 with less training and inference time [23]. YOLOv8 was more accurate on challenging segmentation tasks in industrial surface corrosion detection [24]. In aerial human detection, YOLOv8 performed a bit better in terms of precision and F1-score, whereas the recall was higher in YOLOv5, which is significant to applications where safety is involved [25]. This study recommends using YOLOv5 since it balances precision and performance and provides a real-time solution for durian plantation surveillance on IoT devices like Raspberry Pi.

## 2.3. Proposed System

As highlighted in Table 1, existing systems rely on single algorithms, limited sensors, and less accessible notification platforms, with minimal deterrent mechanisms. A preliminary prototype has been made that incorporate the YOLOv5 and SSD into the Raspberry Pi to evaluate the feasibility of using CNN into an IoT device. SSD has been proven to have inference time of detecting animal faster (in average 10 seconds) than YOLOv5 (in average 27 seconds). However, the accuracy is very low in average 67% compared to YOLOv5 in average 90%. To bridge this gap, the proposed system integrates an IoT-based architecture with CNN models (YOLOv5 and SSD) to achieve higher detection with acceptable speed. A camera sensor provides continuous monitoring, with detected intrusions immediately reported to farmers via Telegram alerts. In addition, an automated sound deterrent (e.g., tiger roar) is activated to repel animals. This hybrid framework ensures real-time monitoring, reliable notifications, and active crop protection, offering a more comprehensive solution for durian plantations.

Table 1. Previous Work Comparison

| Features / Existing systems | Technology | Algorithm | Sensors | | | Notification | | | Deterrent process | |
|---|---|---|---|---|---|---|---|---|---|---|
| | | | Camera | Ultrasonic | UWB-impulse radio | Telegram | Blynk | Message | Sound | Buzzer |
| Yusoff et al., 2022 [3] | Mobile application based, image recognition | Deep Neural Network (DNN) | ✓ | × | × | × | × | × | ✓ | × |
| Panda et al., 2022 [7] | IoT based | - | × | ✓ | × | × | ✓ | × | × | ✓ |
| Sathesh et al., 2022 [5] | Web server, image recognition | YOLO | ✓ | × | × | × | × | ✓ | × | ✓ |
| Anuvind et al., 2023 [8] | Image and Audio processing | YOLO | ✓ | × | × | × | × | ✓ | × | × |
| Xue et al., 2017 [6] | Wireless sensor networks | Convolutional Neural Network (CNN) | × | × | ✓ | × | × | × | × | × |
| Mamat et al., 2022 [9] | IoT based | YOLOv5 | ✓ | × | × | × | × | × | × | × |
| Proposed system | IoT based | YOLOv5, SSD | ✓ | × | × | ✓ | × | × | ✓ | × |

## 3. METHODOLOGY

Figure 1 shows the whole architecture of the animal detection system, including the whole process of the model training and deployment on the Raspberry Pi in real-time. During the development stage, boar, elephant and monkey image datasets (Figure 2) are gathered and pre-processed by the use of annotation and augmentation to subsequently undergo feature extraction. Training is performed twice (on two object detectors, namely, YOLOv5 and SSD) with the respective trained model files (.pt and .tflite, respectively). These trained models can be downloaded and transferred to a network to Raspberry Pi as real-time inference. During the operational stage, the webcam in the system continuously records video frames and using the feature extraction, they are given as input data to the model, which is pre-cached and used in the process of animal recognition. A decision module classifier decides whether to act against a given animal detected. When activated, the system will send a Telegram notification to the plantation owner as well as a deterrent sound to drive the animal away; otherwise nothing happens. This design will provide a unified pipeline consisting of AI-generated detection precision and feasible implementation in a fieldwork environment.

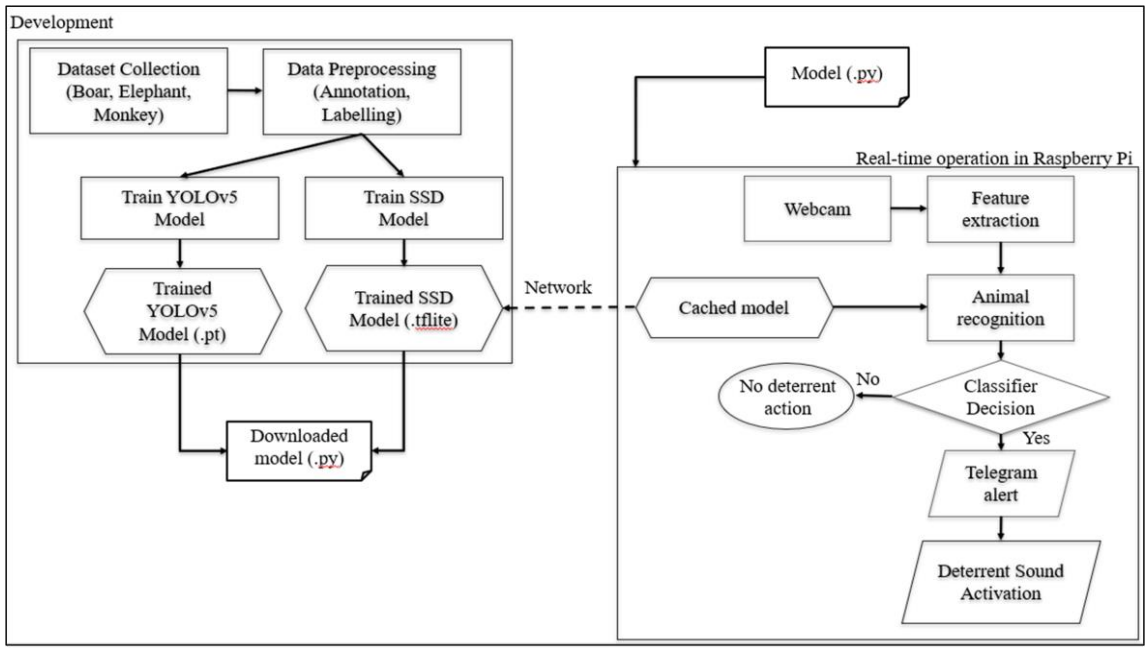

Figure 1. System Architecture of the Animal Detection System from Model Training to Real-Time Deployment on Raspberry Pi.

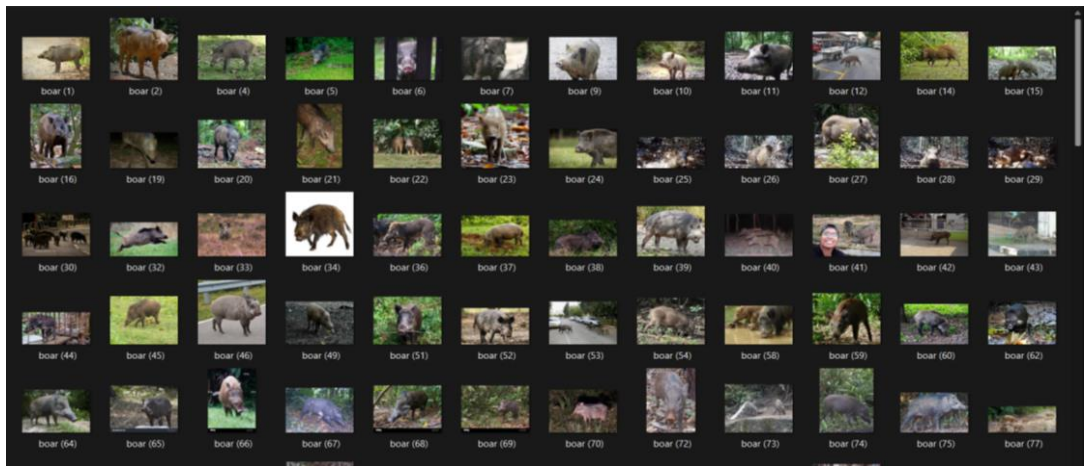

Figure 2. Images of Boar, Elephant and Monkey in various conditions.

### 3.1. Annotation and Augmentation

Before training the model, the dataset needs to be labelled so that the machine learning model is provided with the necessary information about the animal features to learn and make accurate predictions. An open-source tool like Makesense.ai is used to annotate and label the images for training purposes [26]. The labelled image files can then be exported and saved into the 'train' and 'val' folders in the 'labels' folder (refer Figure 3).

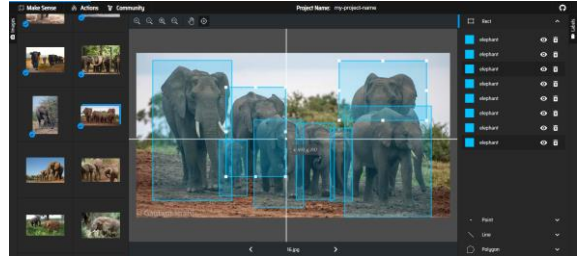

Figure 3. Process of Annotating and Labelling

### 3.2. Train the Model

The model training is done on Google Colab, a free cloud-based platform that allows users to run Python code. The model training starts with installing YOLOv5 and uploading the dataset file into the platform. Once the training is finished, the model will be downloaded and saved into the local host. The same process is repeated for SSD MobileNet model. The model then can be transferred to Raspberry Pi from the local host for full system implementation.

After installing the required dependencies and packages, the train.py can be run for the training process. For this process, a few numbers of epochs have been tried for the model to learn better and give precise predictions. The training process starts with 50 epochs, followed by 150 epochs, and lastly 300 epochs. The reason to start the training process with a low number of epochs and gradually increasing them is to see if there is improvement in the Mean Average Precision (mAP). If there is no improvement in mAP, it is better to stop the training process [27], [28], [29]. The number of epochs must not be too much or too little in order to avoid model from being overfitting or underfitting.

#### 3.2.1. Metrics

This study evaluates model performance using precision (P), recall (R), and mean average precision (mAP). These evaluations are expressed as:

$$P = \frac{TP}{TP + FP} \tag{1}$$

$$R = \frac{TP}{TP + FN} \tag{2}$$

$$AP = \int_0^1 P(R)dR \tag{3}$$

$$mAP = \frac{1}{|Q_R|} \sum_{q=Q_R} AP(q) \tag{4}$$

The average accuracy, denoted as AP, is fundamentally linked to the R index integral to the P index, representing the area under the P-R curve. mAP represents the average accuracy calculated by taking the sum of AP values across all categories and dividing it by the total number of categories.

Table 2. Results of the Model Training

| Class | Images | Instances | Precision | Recall | mAP |
|---|---|---|---|---|---|
| Elephant | 60 | 36 | 0.879 | 0.972 | 0.965 |
| Boar | 60 | 24 | 1 | 0.913 | 0.948 |
| Monkey | 60 | 50 | 0.908 | 0.791 | 0.831 |

Based on the training results (Table 2), it shows that the accuracy of the model in identifying true positives is quite good considering the precision of each class being 0.879, 1, and 0.908. The ability of the model to capture relevant instances is good too due to the recall of each species being 0.972, 0.913, and 0.791 respectively. Overall, the model shows a good performance with the mean average precision (mAP) being over 0.8 for each class. The closer the numbers are to 1, the better the performance.

### 3.3. Combination of a Pre-trained SSD MobileNet Model and YOLOv5 Model

To combine the pre-trained SSD model and the YOLOv5 model, the process starts with downloading the pre-trained model from the Google Colab and saving it in the local host. The same process applies to the YOLOv5 model as well since it was trained on a cloud-based platform. Both models need to be in the same local host. Since the pre-trained model has various classes, the model needs to be filtered so that only specific types of animals can be detected. In this case, the allowed animals to be detected are elephants, boars, and monkeys. Both models need to be loaded and pre-processed to perform prediction. The detected animals with the confidence of 50% and above will be displayed. Figure 4 shows the pseudocode of Load, Pre-process and Prediction using both models.

```
Input:
    V ← video stream from webcam
    M_YOLO ← pretrained YOLOv5 model
    M_SSD ← pretrained SSD MobileNet model
    C ← set of target classes {elephant, boar, monkey}
    T_conf ← confidence threshold (e.g., 0.5)
While webcam is active do:
    f ← capture_frame(V)
    // --- SSD Inference ---
    f_SSD ← preprocess(f, M_SSD)
    D_SSD ← M_SSD.predict(f_SSD)
    For each detection d ∈ D_SSD do:
        If d.confidence > T_conf ∧ d.class ∈ C then:
            draw_bounding_box(f, d)
            Detected_Animals ← Detected_Animals ∪ {d.class}
    // --- YOLOv5 Inference ---
    D_YOLO ← M_YOLO.predict(f)
    For each detection d ∈ D_YOLO do:
        If d.confidence > T_conf ∧ d.class ∈ C then:
            draw_bounding_box(f, d)
            Detected_Animals ← Detected_Animals ∪ {d.class}
```

Figure 4. Pseudocode of Combining Both Model

### 3.4. Deterrence Module using Telegram Bot

The animal detection system is programmed so that when an animal is detected, it will capture the image of the detected animal and send it to the Telegram bot along with an alert message. Upon receiving the message, users can reply to activate and stop the deterrent sound by using 'deter' and 'stop' commands.

## 4. RESULTS AND DISCUSSION

To understand the efficiency of the animal detection system, a set of structured experiments were taken to examine the performance of the system under various environmental conditions and possible interaction schemes. The test consisted in five basic components: accuracy of recognition of animals, enabling the sound of deterrence, combining Telegram bots, management of distal access, a comparison of detection models. The tests were dependent on the concerned performance measure like detection accuracy, inference time, alert responsive and command success rate.

### 4.1. Model Performance Test

The testing dataset consists of 10 images each for boar, elephant and monkeys. The chosen images are of different environmental circumstances in order to get varied testing results. The SSD, YOLOv5 and combined SSD+YOLOv5 were evaluated to determine the most suitable configuration of the system. Based on the model performance results in Table 4-5 and Figure 5-6, the SSD model has a low average accuracy, while the inference time is shorter. The YOLO model has a high average of accuracy, while the time taken to detect the animals are longer. On the other hand, the combination model has a high accuracy level, but the inference time is slightly shorter than the YOLO model.

Table 4. Average Accuracy (%) of Each Model Across Three Animal Species

|  | Boar | Elephant | Monkey |
|---|---|---|---|
| **SSD** | 65.7 | 71 | 62.2 |
| **YOLOv5** | 92.4 | 89.8 | 80.3 |
| **SSD+YOLOv5** | 90 | 91.6 | 82.2 |

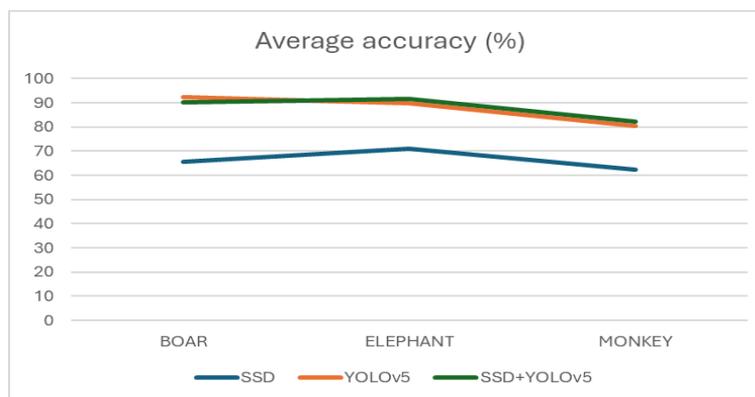

Figure 5. Comparison of Average Accuracy for SSD, YOLOv5, and YOLO+SSD Models in Boar, Elephant, and Monkey Detection

Table 5. Inference Time (s) of Each Model Across Three Animal Species

|  | Boar | Elephant | Monkey |
|---|---|---|---|
| **SSD** | 12.7 | 9.3 | 14 |
| **YOLOv5** | 29 | 27.8 | 26.3 |
| **SSD+YOLOv5** | 20.8 | 22 | 24.3 |

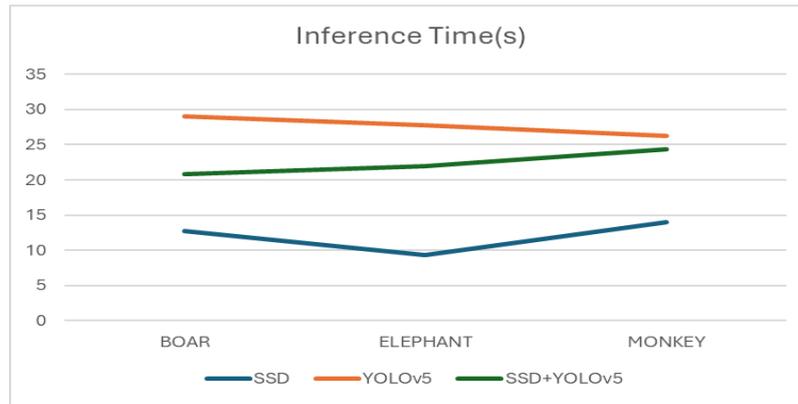

Figure 6. Comparison of Inference Time(s) for SSD, YOLOv5, and YOLO+SSD Models in Boar, Elephant, and Monkey Detection

## 4.2. Animal Recognition (Day)

Figure 7 and Table 6 shows that the animal species boar, elephant, and monkey are accurately recognized during the day. From the results, it can be concluded that the recognition results do not show much difference between still pictures and live videos in terms of accuracy level except in the aspect of speed which takes a slightly longer time to detect.

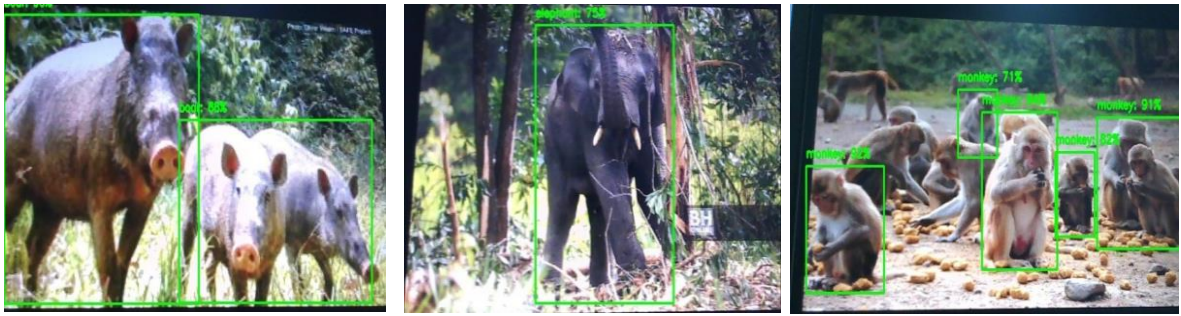

Figure 7. Result of animal recognition (from left to right: Boar, Elephant and Monkey)

Table 6. Results of Animal Recognition on Daylight

| Class | Still Picture | | Video | |
|---|---|---|---|---|
|  | Average Accuracy (%) | Average Time Taken (s) | Average Accuracy (%) | Average Time Taken (s) |
| Boar | 91.3 | 26 | 90.8 | 26.5 |
| Elephant | 88.2 | 24.6 | 88.6 | 26 |
| Monkey | 80.3 | 20.6 | 80.1 | 23.1 |

### 4.3. Animal Recognition (Night)

Figure 8 and Table 7 shows that the animal species boar, elephant, and monkey are accurately recognized during the night as well. From the results, the accuracy level is slightly lower than during the day. The time taken is also slightly longer in order to get an accurate result. This shows that the bad lighting may affect the performance of the system. It can be concluded that the recognition results for live videos take a longer time to detect the animals at night.

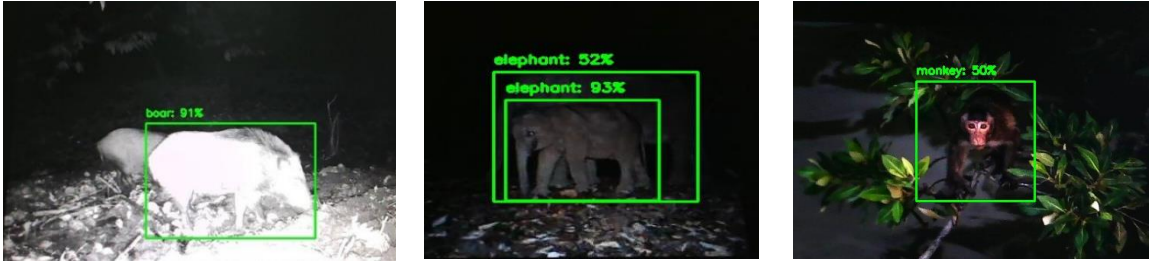

Figure 8. Result of animal recognition (from left to right: Boar, Elephant and Monkey)

Table 7. Results of Animal Recognition (Night)

| Class | Still Picture | | Video | |
|---|---|---|---|---|
| | Average Accuracy (%) | Average Time Taken (s) | Average Accuracy (%) | Average Time Taken (s) |
| Boar | 83.8 | 19.4 | 80.2 | 24 |
| Elephant | 79.2 | 24 | 79.3 | 25.3 |
| Monkey | 87.3 | 25 | 85.7 | 27 |

### 4.4. Confusion Matrix for Animal Recognition

To determine the confusion matrix, a total amount of 20 tests have been done for each animal species. Based on the results in Figure 9 and Table 8, the system was able to accurately recognize the animal species 85% of the time for Boar, 90% of the time for Elephant, and 70% of the time for Monkey. The remaining percentage that contributed to the false positive and false negative outcomes are the results from bad lighting and other environmental circumstances such as trees that were blocking the camera view as well as the distance between the camera and the animals. The angle of the animal's body plays a role too in determining an accurate recognition.

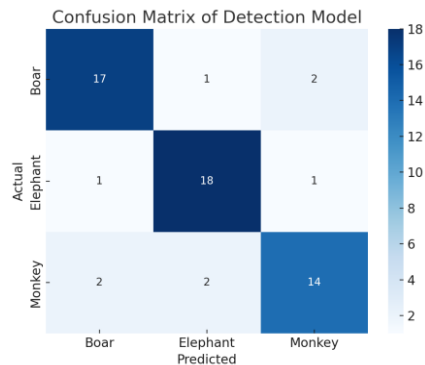

Figure 9. Confusion Matrix for SSD+YOLOv5 Model

## 4.5. Model Detection Limitations

It should be noted that the dataset size used in this study was relatively small (approximately 60 training images and 10 testing images per class). While the hybrid YOLOv5–SSD model achieved promising accuracy, the limited dataset may restrict the generalizability of the findings. Future work will require the use of larger, more diverse datasets to enhance robustness across different environments. In addition, inference time measurements were reported based on single-run results; although consistent trends were observed, a more rigorous statistical evaluation (e.g., multiple test runs with variance analysis) will be conducted in future studies to ensure robustness and reliability of inference performance.

## 4.6. System Integration

Figure 10 shows the hardware setup between Raspberry Pi 5, webcam and external speaker. When an animal is detected in the plantation area, the system will alert users by sending a message on Telegram (Figure 11). Users then can activate the deterrent sound by using the command 'deter'. The system will play the deterrent sound on loop. Once there is no animal detected in the area anymore, users can stop the deterrent sound by using the command 'stop' (Figure 12). The deterrent sound consists of the sound of a tiger's roar to scare the animals away from the plantation area.

Table 8. Performance Metric for Animal Recognition

| Class | Accuracy | Precision | Recall | F1-score |
|-------|----------|-----------|--------|----------|
| Boar | 0.85 | 0.89 | 0.94 | 0.92 |
| Elephant | 0.90 | 0.95 | 0.95 | 0.95 |
| Monkey | 0.70 | 0.88 | 0.78 | 0.82 |

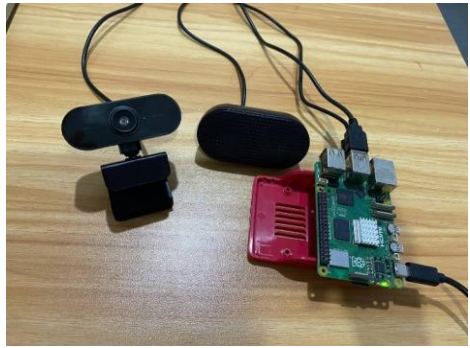

Figure 10. Hardware setup for Animal Detection System

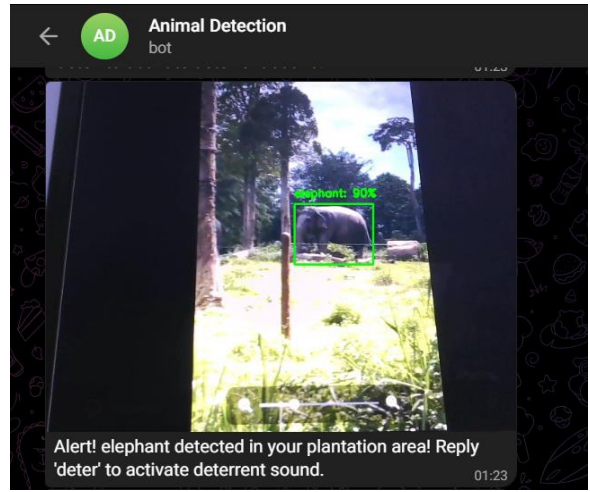

Figure 11. Alert notification sent to Telegram

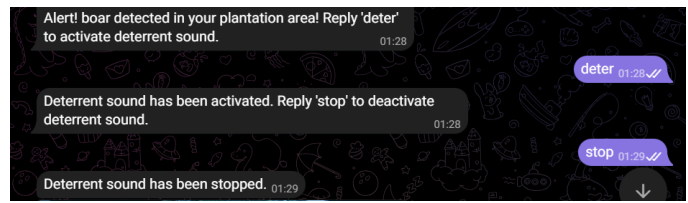

Figure 12. The deterrent sound activation process

# 5. CONCLUSION AND FUTURE WORK

In conclusion, the proposed hybrid YOLOv5–SSD system demonstrates a balanced trade-off between detection accuracy and inference speed, making it practical for deployment on resource-constrained IoT devices such as Raspberry Pi. By integrating real-time alerts through Telegram and a deterrent mechanism, the system provides farmers with immediate and automated responses to animal intrusions, reducing crop damage without requiring constant human presence. These contributions directly address real-world plantation challenges by enhancing surveillance reliability, lowering labor costs, and mitigating financial losses caused by wildlife. While the current implementation is limited by dataset size and environmental conditions such as low light, the study lays a foundation for scalable smart farming solutions. Future enhancements will focus on expanding datasets, improving low-light detection, and integrating additional IoT sensors to increase robustness and deployment feasibility in large-scale plantations.

## ACKNOWLEDGEMENTS

The authors would like to thank the Center for Advanced Computing Technology (C-ACT), Fakulti Teknologi Maklumat dan Komunikasi, Universiti Teknikal Malaysia Melaka (UTeM) for supporting financially the work done in this paper.

# AUTHORS

**Anis Suttan Shahrir** holds her Diploma in Computer Science from Universiti Malaysia Pahang (UMP), and a Bachelor's of Computer Science in Computer Networking from Universiti Teknikal Malaysia Melaka (UTeM), Malaysia. She is currently a System Engineer at Mesiniaga Berhad. Her research interests include computer vision, IoT-based detection systems, automation, and system reliability. You can contact her at email anisshahrirr@gmail.com.

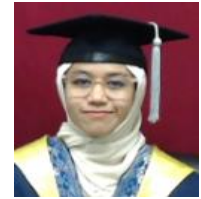

**Zakiah Ayop** graduated with Bachelor of Computer Science and Master of Science in Distributed Computing from Universiti Teknologi Malaysia and Universiti Putra Malaysia respectively. Currently, she is a senior lecturer in Department of Computer System and Communications, Faculty of Information and Communication Technology (FTMK), UTeM. Her research interest is Information System, Internet of Things (IoT), Cyber-physical System (CPS) and Networking. She is a certified CCNA Instructor. Her work has been presented at various international conferences and published in journals, contributing significantly to advancements in her fields of interest.

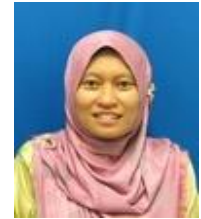

**Syarulnaziah Anawar** holds her Bachelor of Information Technology from UUM, Msc in Computer Science from UPM, and PhD in Computer Science from UiTM, Malaysia. She is currently a Senior Lecturer at the Department of Computer and Communication System, Faculty of Information and Communication Technology, UTeM. She is a member of the Information Security, Digital Forensic, and Computer Networking (INSFORNET) research group. Her research interests include human-centered computing, participatory sensing, mobile health, usable security and privacy, and societal impact of IoT.

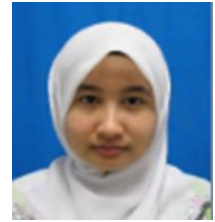

**Norulzahrah Mohd Zainuddin** received the M.Sc. degree from Universiti Putra Malaysia and joined Military Academy of Malaysia, in 2002. She is currently a Lecturer with the Department of Computer Science, Faculty of Defence Science and Technology, National Defence University Malaysia (UPNM). She is also a member of the Informatics Intelligence Special Interest Group, UPNM. She has published a number of papers in international journals and conferences. Her main research interests include forensic computing, online social networks, and computer intelligence.

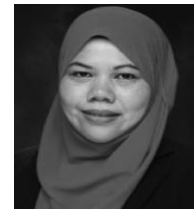